%% file: sample-sigconf.tex
\documentclass[sigconf]{acmart}
\AtBeginDocument{%
  }

\copyrightyear{2026}
\acmYear{2026}
\setcopyright{cc}
\setcctype{by-nc-nd}
\acmConference[KDD 2026] {Proceedings of the 32nd ACM SIGKDD Conference on Knowledge Discovery and Data Mining V.2}{August 9--13, 2026}{Jeju Island, Republic of Korea.}
\acmBooktitle{Proceedings of the 32nd ACM SIGKDD Conference on Knowledge Discovery and Data Mining V.2 (KDD 2026), August 9--13, 2026, Jeju Island, Republic of Korea}
\acmISBN{979-8-4007-2259-2/2026/08}
\acmDOI{10.1145/3770855.3817889}

\usepackage{bm}
\usepackage{mathtools}
\usepackage{subcaption}
\usepackage[nopar]{lipsum}
\usepackage{amsmath}

\usepackage{amssymb}
\usepackage{cleveref}
\usepackage{array}
\usepackage{multirow}
\usepackage[table]{xcolor}
\usepackage{wrapfig}
\usepackage{marvosym}
\usepackage{enumitem}
\usepackage{algorithm}
\usepackage{multirow}
\usepackage{float}
\usepackage{algpseudocode}

\definecolor{mgreen}{RGB}{255, 133, 82}
\definecolor{mgreen_vis}{RGB}{0, 113, 45}
\definecolor{mprior_vis}{RGB}{52, 117, 172}
\definecolor{mredred}{RGB}{184, 0, 31}
\definecolor{loss_red}{RGB}{196, 39, 39}
\definecolor{mlblue}{RGB}{124, 183, 201}
\definecolor{mblue}{RGB}{34, 199, 218}
\definecolor{mgreenpp}{RGB}{95, 139, 76}
\definecolor{mred}{RGB}{157, 33, 34}
\definecolor{myellow}{RGB}{237, 190, 67}
\definecolor{mgreenpp}{RGB}{52, 121, 40}
\definecolor{mgreen_vis}{RGB}{0, 113, 45}
\definecolor{mprior_vis}{RGB}{74, 98, 138}
\definecolor{mbevlane}{RGB}{230, 197, 72}
\definecolor{morange}{RGB}{34, 199, 218}
\definecolor{fig_per}{RGB}{164, 128, 41}
\definecolor{fig_rea}{RGB}{94, 138, 152}
\definecolor{fig_sup}{RGB}{136, 75, 120}
\definecolor{rev}{RGB}{191, 26, 26}
\definecolor{tab_per}{RGB}{243, 241, 232}
\definecolor{tab_rea}{RGB}{217, 228, 232}
\definecolor{tab_sup}{RGB}{226, 214, 224}
\definecolor{grayres}{RGB}{38, 38, 38}

\newcommand{\bzr}{B\'{e}zier\ }
\newcommand{\modelname}{CoPo\ }

\newcommand{\curve}{curve-guided}
\newcommand{\Curve}{Curve-Guided}

\newcommand{\eg}{\emph{e.g.}}
\newcommand{\ie}{\emph{i.e.}}

\crefname{figure}{Fig.}{Figs.}
\Crefname{figure}{Figure.}{Figures.}
\crefname{table}{Tab.}{Tabs.}
\Crefname{table}{Table.}{Tables.}
\crefname{section}{Sec.}{Secs.}
\Crefname{section}{Section.}{Sections.}
\crefname{equation}{Eq.}{Eqs.}

\begin{document}

\title{Geometry-Guided Representations for Coherent Lane and Traffic Topology Reasoning in Driving Scenes}
\author{Yueru Luo}
\email{222010057@link.cuhk.edu.cn}
\orcid{0000-0002-7684-3229}
\affiliation{%
  \institution{Shenzhen Future Network of Intelligence Institute}
  \city{Shenzhen}
  \state{Guangdong}
  \country{China}
}
\affiliation{%
  \institution{School of Science and Engineering, The Chinese University of Hong Kong, Shenzhen}
  \city{Shenzhen}
  \state{Guangdong}
  \country{China}
}

\author{Changqing Zhou}
\email{czhou149@connect.hkust-gz.edu.cn}
\orcid{0000-0002-7269-6380}
\affiliation{%
  \institution{The Hong Kong University of Science and Technology (Guangzhou)}
  \city{Guangzhou}
  \state{Guangdong}
  \country{China}
  }

\author{Yiming Yang}
\email{224010097@link.cuhk.edu.cn}
\orcid{0009-0005-7466-1041}
\affiliation{
  \institution{Shenzhen Future Network of Intelligence Institute}
  \city{Shenzhen}
  \state{Guangdong}
  \country{China}
}
\affiliation{%
  \institution{School of Science and Engineering, The Chinese University of Hong Kong, Shenzhen}
  \city{Shenzhen}
  \state{Guangdong}
  \country{China}
}

\author{Erlong Li}
\email{erlongli@tencent.com}
\orcid{0009-0005-1008-8311}
\author{Chao Zheng}
\correspondingauthor
% \,\textsuperscript{\Letter}}
\email{chrisczheng@tencent.com}
\orcid{0009-0006-6283-6980}
\affiliation{%
  \institution{Tencent T Lab}
  \city{Beijing}
  \country{China}}

\author{Shuguang Cui}
\email{shuguangcui@cuhk.edu.cn}
\orcid{0000-0003-2608-775X}
\affiliation{%
  \institution{School of Science and Engineering, The Chinese University of Hong Kong, Shenzhen}
  \city{Shenzhen}
  \state{Guangdong}
  \country{China}
}
\affiliation{%
  \institution{Shenzhen Future Network of Intelligence Institute}
  \city{Shenzhen}
  \state{Guangdong}
  \country{China}
}

\author{Zhen Li}
\correspondingauthor
\email{lizhen@cuhk.edu.cn}
\orcid{0000-0002-7669-2686}
\affiliation{%
  \institution{School of Science and Engineering, The Chinese University of Hong Kong, Shenzhen}
  \city{Shenzhen}
  \state{Guangdong}
  \country{China}
}
\affiliation{%
  \institution{Shenzhen Future Network of Intelligence Institute}
  \city{Shenzhen}
  \state{Guangdong}
  \country{China}
}

% \thanks{\Letter\, denotes corresponding authors.}
% \correspondingauthorfootnote{denotes corresponding authors.}
\renewcommand{\shortauthors}{Yueru Luo et al.}

\begin{abstract}
  Road topology reasoning is fundamental for autonomous driving, requiring both accurate perception of road elements and understanding of their complex connectivity, including lane connectivity (Lane-to-Lane, L2L) and traffic regulation (Lane-to-Traffic signs, L2T). However, existing methods typically treat perception and topology reasoning as fragmented tasks, ignoring their potential for mutual enhancement. Crucially, while topology is inherently relational, prior works often overlook geometric relationships during feature extraction, relying instead on brittle post-processing or coordinate-based heuristics applied only at inference time. To bridge this gap, we propose \textbf{CoPo} (\textbf{Co}herent Perception and to\textbf{Po}logy), a unified framework that
  integrates geometry-guided relational modeling across three levels:
  \textbf{1)} Perception-level: We introduce a relation-aware lane detector that utilizes geometry-biased self-attention and curve-guided cross-attention to enrich lane representations with structural priors; \textbf{2)} Reasoning-level: We design relation-enhanced topology heads, including a geometry-enhanced L2L head and a cross-view L2T head, which effectively align features to infer connectivity; and \textbf{3)} Supervision-level: We implement a contrastive InfoNCE strategy to regularize relational embeddings, pulling connected pairs closer in the latent space.
  This coherent multi-level design enables end-to-end joint optimization of perception and reasoning.
  Extensive experiments on OpenLane-V2 demonstrate that \modelname significantly outperforms existing methods, achieving gains of {\textbf{+3.1}} in DET$_l$, {\textbf{+5.3}} in TOP$_{ll}$, {\textbf{+4.9}} in TOP$_{lt}$, and {\textbf{+4.4}} overall in OLS, setting a new state-of-the-art.
\end{abstract}



\begin{teaserfigure}
\centering
  \includegraphics[width=0.96\textwidth]{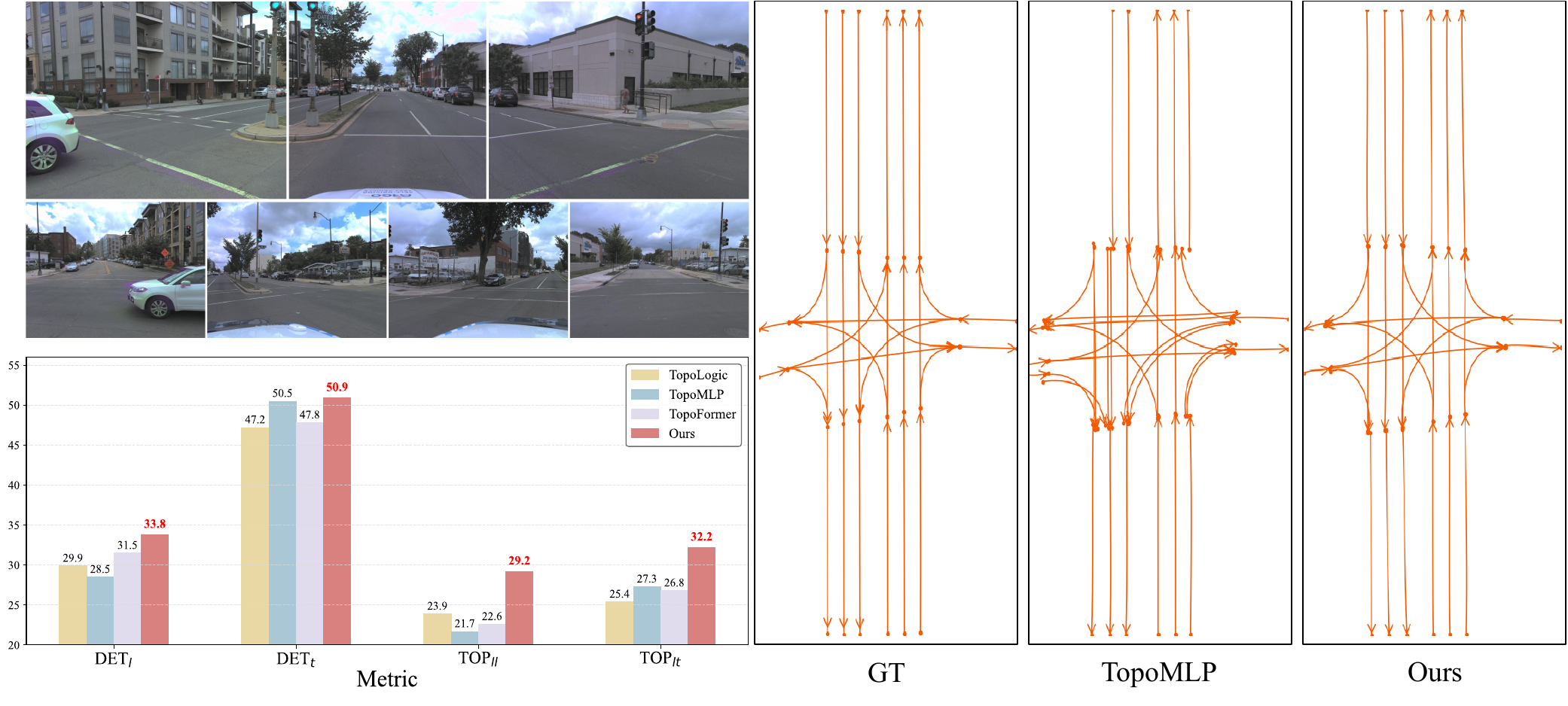}
  \vspace{-4mm}
  \caption{\textbf{Performance and visualization comparison between previous methods and ours.} The top-left shows multi-view input images. Our approach, \textcolor{mred}{\modelname}, integrates relational modeling across multiple levels, strengthening both perception and topology reasoning. As shown in the bottom-left quantitative results on OpenLane-V2, \modelname significantly outperforms prior methods across all metrics. On the right, qualitative comparisons demonstrate that {\modelname} produces more accurate lane geometries and well-aligned connectivity than prior approaches. 
  }
  \Description{Performance and visualization comparison between previous methods and ours.}
  \label{fig:teaser}
\end{teaserfigure}

\maketitle

\newcommand\kddavailabilityurl{https://doi.org/10.5281/zenodo.20455469}
\ifdefempty{\kddavailabilityurl}{}{
\begingroup\small\noindent\raggedright\textbf{KDD Availability Link:}\\
The source code and data of this paper has been made publicly available at \url{\kddavailabilityurl}.
\endgroup
}

\input{sections/1_intro}
\input{sections/2_related}

\input{sections/3_method}

\input{sections/4_exp}

\clearpage
\bibliographystyle{ACM-Reference-Format}
\bibliography{sample-base}

\appendix

\section{Implementation Details}
\label{sec:supp_imple}

\subsection{Overall Optimization Objective}  
\label{sec:supp_loss}
As discussed in the main paper, our training objective comprises detection losses for individual elements and reasoning losses for their relationships.
Following prior methods~\cite{toponet,topomlp,deformdetr}, traffic element detection is supervised using a combination of classification loss and regression loss.
Lane detection is similarly optimized with classification and regression losses. The regression loss includes L1 loss on Bézier control points and a chamfer distance loss $L_{\text{B\'{e}zierCD}}$ computed on sampled on-curve points (introduced in our curve-guided cross-attention). We sample $K=11$ points along each Bézier curve.
Detection loss weights follow previous works~\cite{topomlp}.
For topology reasoning, we adopt focal loss for relation classification and introduce a contrastive loss $\mathcal{L}_{\text{con}}$ to enhance relational learning in the embedding space. Each contrastive pair includes one positive and three negative samples. The focal and contrastive losses are weighted by $\lambda_{1} = 5$ and $\lambda_{2} = 0.1$, respectively.
The complete loss function is defined as:
\begin{align}
    \mathcal{L}_{\text{all}} &= \mathcal{L}_{\text{det}} + \mathcal{L}_{\text{relation}}, \\
    \mathcal{L}_{\text{relation}} &= \lambda_{1} \mathcal{L}_{\text{class}}^{\text{topo}} + \lambda_{2} \mathcal{L}_{\text{con}}.
\end{align}

\section{More Qualitative Results}
\label{sec:supp_qua_res}
\begin{figure*}[htp]
    \centering
    \includegraphics[width=0.82\linewidth]{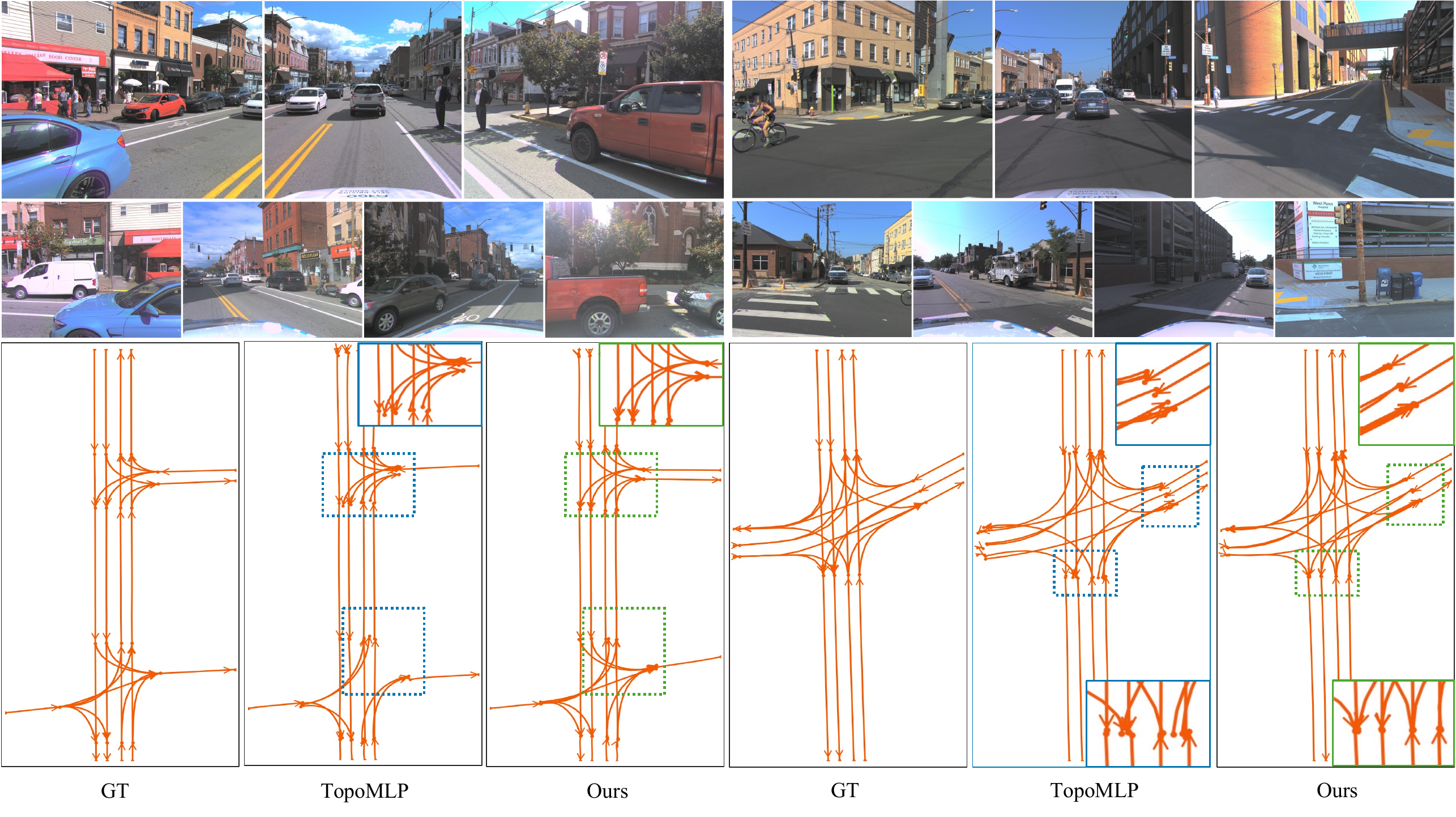}
    \caption{
    Comparative visual results on OpenLane-V2 \texttt{subsetA}. 
    The top row shows multi-view input images, the bottom row shows lane predictions.
    We show comparison between ground truth, TopoMLP~\cite{topomlp} and ours. %
    The {blue} box highlights misaligned connection point predictions from {TopoMLP}, and the \textcolor{mgreenpp}{\textbf{green}} box shows the corresponding aligned predictions from \textcolor{mgreenpp}{our \textbf{\modelname}}.
    For clarity, zoomed-in views of selected regions are displayed at the top-right or bottom-right corners.
    }
    \label{fig:vis_comp}
\end{figure*}

We present additional qualitative results in~\cref{fig:vis_comp} (regarding the lane centerline predictions) and~\cref{fig:vissupp2} (covering predictions: lane centerlines, L2L and L2T topology relationship estimation). Leveraging our proposed components, \modelname demonstrates superior perception of lane centerlines compared to previous methods, as shown in~\cref{fig:vis_comp} and the second row of the figures. Moreover, \modelname achieves more accurate lane-to-lane (L2L) and lane-to-traffic-element (L2T) topology reasoning, as illustrated in the last two rows. Notably, \modelname exhibits significantly improved lane perception accuracy and topology reasoning performance in complex scenarios involving intricate L2T relationships.  

\section{Extra Experiments}

\subsection{More Analysis on Geometry Topology.}
\label{sec:supp_exp_topologic}
As discussed in the main paper, TopoLogic~\cite{topologic} proposes a geometric distance topology (GDT), 
which shares certain similarities with our Geometry-Biased Self-Attention and Geometry-Enhanced L2L Topology, but \textit{differs fundamentally in design and effectiveness}. Below we provided more detailed analysis. 

\begin{table}[h!]
\centering
\caption{\textbf{Experiments on Geometric Distance Topology (GDT).} ``Topologic -- {GDT}$_{\text{L2L}}$'': Inference without GDT$_{\text{L2L}}$. {TopoLogic}$^\dagger$: Inference using the hyperparameters learned during their lane representation training. {``Ours + {GDT}$_{\text{lane}}$''}: Replaces our geometry encoding with TopoLogic’s GDT for lane representation learning. {``Ours + {GDT}$_{\text{L2L}}$''}: Ensembles our L2L predictions with GDT, same as TopoLogic.}
\label{tab:compare_logic_supp}
\renewcommand{\arraystretch}{0.9}
\begin{tabular}{l|cccc|c}
    \toprule
    \noalign{\vskip -0.3mm}
     Model & DET$_l$ & DET$_{t}$ & TOP$_{ll}$ & TOP$_{lt}$ & OLS \\
    \hline
    \noalign{\smallskip}
    Topologic & 29.9 & 47.2 & 23.9 & 25.4 & 44.1 \\
    Topologic -- {GDT}$_{\text{L2L}}$ & 29.9 & 47.2 & 11.6 & 25.4 & 40.4 \\
    Topologic$^\dagger$ & 29.9 & 47.2 & 23.2 & 25.4 & 43.9 \\
    \hline
    \noalign{\smallskip}
    Ours & \textbf{33.8} & \textbf{50.9} & \textbf{29.2} & \textbf{32.2} & \textbf{48.9} \\
    Ours + {GDT}$_{\text{lane}}$ & 32.8 & 50.0 & 28.4 & 30.8 & 47.9 \\
    Ours + {GDT}$_{\text{L2L}}$ &33.8 & 50.9 & 28.7 & 32.2 & 48.7 \\
    \bottomrule
\end{tabular}
\end{table}

\paragraph{\textbf{Lane Representation Learning.} GDT~\cite{topologic,topoformer} focuses exclusively on connectivity estimation by computing the end-to-start point distance between lanes and updating representations using an additional GCN block.
\textit{In contrast}, our method captures richer spatial cues beyond connectivity, such as inter-lane distances and angular relations. These geometric priors—\eg, parallelism, perpendicularity, and merging—are common in real-world road layouts and intuitively leveraged by humans. We explicitly encode the shortest endpoint distance and angular difference between lanes into high-dimensional relational features, which are further projected as biases in self-attention to enhance both perception and downstream topology reasoning.}

\paragraph{\textbf{L2L Topology Reasoning.} Our method leverages the observation that annotated connected lanes share overlapping endpoints. We encode end-to-start distances as high-dimensional embeddings and integrate them directly into L2L relation features during training, allowing the model to learn geometric cues end-to-end. In contrast, TopoLogic~\cite{topologic} computes a scalar topology score for each lane pair based on GDT and applies it to adjust L2L predictions ({GDT}$_\text{L2L}$) \textbf{only} at inference time. This post-hoc refinement is excluded from learning and relies on manually assigned hyperparameters, which may not generalize across scenarios. As shown in~\cref{tab:compare_logic_supp}, this design limits performance and robustness, further highlighting the advantage of our approach.}

To better understand these differences, we conduct three comparative experiments in~\cref{tab:compare_logic_supp} and analyze them below.

\noindent\textbf{Exp. 1: GDT \textit{vs.} Geometry-Biased Self-Attention.} 
We replace our geometry encoding with TopoLogic's end-to-start distance to construct the self-attention bias for lane representation learning. As shown
in~\cref{tab:compare_logic_supp}, this modification degrades DET$_l$ by 1.0, TOP$_{ll}$ by 0.8, and TOP$_{lt}$ by 1.4, suggesting that GDT is insufficient for capturing rich inter-lane spatial relationships and reaffirming the advantage of our relation embedding.

\noindent\textbf{Exp. 2: Evaluating GDT in L2L Topology Reasoning.}
Unlike our geometry-enhanced L2L topology head, which integrates geometric cues into relation embeddings during training, TopoLogic applies GDT only as an inference-time adjustment, using manually assigned hyperparameters to balance GDT and model-based predictions rather than the weights learned during lane representation training. We evaluate TopoLogic's model under two variants: TopoLogic$^\dagger$, which uses the learned hyperparameters, and TopoLogic -- GDT$_\text{L2L}$, which removes GDT
from L2L inference. As shown in~\cref{tab:compare_logic_supp}, both variants degrade performance, with a TOP$_{ll}$ drop when GDT is removed, suggesting that TopoLogic's lane features alone do not sufficiently encode relational information and that post-hoc/manual adjustment limits robustness and generalizability.

\noindent\textbf{Exp 3: Applying GDT to Our L2L Inference.}
We further investigate the generalizability of GDT by incorporating it into our model’s L2L inference, following TopoLogic’s ensembling strategy. Specifically, we fuse our L2L predictions with GDT outputs using their predefined hyperparameters, without modifying our trained model. As shown in~\cref{tab:compare_logic_supp}, this ensembling leads to a decrease in TOP$_{ll}$ by 0.5 (Ours + {GDT}$_\text{L2L}$ \textit{vs.} Ours).
This outcome highlights the limited transferability of TopoLogic’s formulation. In contrast, our model achieves strong L2L performance without requiring additional tuning or post-processing, demonstrating that our relation embedding offers a more effective and learnable approach to modeling geometric topology.

\section{Limitations and Future Work}
  \label{sec:supp_failcase}
  Through qualitative analysis, we identify recurring challenging cases for topology reasoning, including small traffic elements, distant lanes, and occluded or partially
  visible lanes. Since OpenLane-V2 does not provide explicit scenario annotations such as occlusion labels, we further conduct attribute-based quantitative breakdowns for lane
  distance and traffic-element size.

  \begin{table}[h]
  \centering
  \small
  \begin{minipage}[t]{0.47\linewidth}
  \centering
  \caption{Lane detection by distance. \%GT denotes the percentage of total lane instances.}
  \label{tab:failure_lane_distance}
  \setlength{\tabcolsep}{5pt}
  \renewcommand{\arraystretch}{0.9}
  \begin{tabular}{lcc}
  \toprule
  Distance & DET$_l$ & \%GT \\
  \midrule
  Near ($<25$m) & 34.4 & 45.3\% \\
  Far ($\ge25$m) & 30.2 & 54.7\% \\
  \bottomrule
  \end{tabular}
  \end{minipage}
  \hfill
  \begin{minipage}[t]{0.47\linewidth}
  \centering
  \caption{TE detection by size. Area ratio denotes element area divided by image area.}
  \label{tab:failure_te_size}
  \setlength{\tabcolsep}{5pt}
  \renewcommand{\arraystretch}{0.9}
  \begin{tabular}{lcc}
  \toprule
  Area ratio & DET$_t$ & \%GT \\
  \midrule
  $<0.01\%$ & 13.2 & 19.2\% \\
  $<0.05\%$ & 41.5 & 68.7\% \\
  $\ge0.05\%$ & 59.9 & 31.3\% \\
  \bottomrule
  \end{tabular}
  \end{minipage}
  \end{table}

  As shown in~\cref{tab:failure_lane_distance,tab:failure_te_size}, far-range lanes and small traffic elements are the dominant challenges. Far lanes account for 54.7\% of
  lane instances and obtain lower DET$_l$ than near lanes. Similarly, small traffic elements are much harder to detect: elements with area ratio below 0.01\% only achieve 13.2
  DET$_t$, while larger elements ($\ge0.05\%$) reach 59.9 DET$_t$. These detection failures can further affect downstream lane-to-lane and lane-to-traffic topology reasoning.

  These challenging cases point toward future directions: incorporating temporal continuity across frames, leveraging motion cues or sequential context, and fusing richer
  spatiotemporal relational information may help improve robustness in far-range, small-object, and occluded-scene cases.

\end{document}

%% file: sections/1_intro.tex
\section{Introduction}
\label{sec:intro}

Understanding road topology~\cite{traffickdd} is fundamental for autonomous driving, as it provides structured knowledge of lanes and traffic elements for navigation and compliance with traffic rules. A complete topology model requires not only perceive road elements but also reason about how they connect: how lanes interconnect (Lane-to-Lane, L2L), and how traffic elements (\eg, lights or signs) regulate specific lanes (Lane-to-Traffic-element, L2T).
We follow the OpenLane-V2 benchmark~\cite{Openlane-v2}, where \textbf{perception} involves detecting lanes in 3D space from multi-view (MV) images and recognizing traffic elements in the front-view (FV), while \textbf{reasoning} infers L2L connectivity among lanes and L2T associations between lanes and traffic elements within only the FV.

Current approaches~\cite{toponet, topomlp, topologic,roadpainter} typically organize topology reasoning into a sequential perception-then-reasoning pipeline.
Despite recent advances, two major gaps remain.
\textbf{First, fragmented task optimization:} 
methods often prioritize either perception or L2L reasoning, leaving L2T reasoning underexplored and seldom optimizing these components jointly.
This fragmented treatment overlooks the opportunity for mutual enhancement between perception and reasoning, as the lack of end-to-end relational coupling prevents effective across-stage synergy, thereby constraining holistic optimization.
Concretely,
in \textbf{perception}, 
many works~\cite{toponet,topomlp,roadpainter} adapt generic object detectors~\cite{petr,deformdetr} by converting bounding boxes into lane points or curves, but overlook intrinsic geometric relationships~\cite{li2022reconstruct,gemap} (\eg, parallelism, continuity) that humans naturally exploit.
TopoLogic~\cite{topologic} makes a partial attempt by encoding end-to-start distances as a topology prior via a GNN, yet this design is narrowly tailored to connectivity and relies on a separate graph module. A key question remains: \textit{how can we inject structural cues directly into perception so that lane features become inherently relation-aware?}
In \textbf{reasoning}, existing works rely heavily on coordinate encodings that are brittle to small endpoint shifts, making L2L prediction fragile. 
TopoLogic~\cite{topologic} partly mitigates this issue via geometric distance in post-processing, but our study (\textit{\cref{tab:compare_logic} and~\cref{sec:supp_exp_topologic}}) shows that performance degrades sharply when removing that post-processing, implying that the model fails to internalize relational modeling.
L2T reasoning is even less explored: most methods naïvely fuse BEV lane features and FV traffic elements, ignoring cross-view misalignment. Some approaches~\cite{topo2d} mitigates this issue by leveraging 2D lane features from an additional decoder, but adds computational overhead.

\textbf{Second, weak relational modeling:} Although topology reasoning inherently involves relation prediction, most works directly force the model relation prediction via supervision~\cite{toponet,topomlp,topo2d,roadnet} or post-processing (\eg, distance-based scoring in TopoLogic~\cite{topologic}), rather than embedding relational cues into feature learning and connectivity prediction. As a result, structural priors (such as parallelism or proximity) between lanes and traffic elements are underutilized, leaving models brittle to geometric variation.

To address these gaps, we propose \modelname, a \textbf{multi-level relational modeling} approach that that systematically integrates relational modeling across three levels. This enables perception and reasoning to be optimized jointly and coherently, grounded by structural relationships inherent in road scenes:
\textbf{1) Relational Perception:} we develop a relation-aware lane detector based on a compact Bézier-curve representation. A \textit{Geometry-Biased Self-Attention} module encodes inter-lane affinities (distance, orientation) as attention biases, while a \textit{Curve-Guided Cross-Attention} module aggregates contextual cues along curves, ensuring relational awareness even with sparse control points.  
\textbf{2) Relational Reasoning:} we design relation-enhanced topology heads: a \textit{Geometry-Enhanced L2L head} that embeds inter-lane distances to produce robust connectivity predictions, and a \textit{Cross-View L2T head} that aligns BEV lanes with FV traffic elements to overcome spatial discrepancy.
\textbf{3) Relational Supervision:} we introduce a InfoNCE~\cite{wu2018unsupervised,He2020moco,liu2021self}-based contrastive objective that pulls connected pairs closer and separates non-connected ones in the embedding space, regularizing relational learning.

Overall, our contributions can be summarized as follows:

\begin{itemize}[leftmargin=3.8mm]
\vspace{-2pt}
    \item We identify two core limitations in existing topology reasoning, namely fragmented task optimization and weak relational modeling, and highlight the need for a unified relational perspective.
    \item  We propose \modelname, a multi-level relational modeling approach that integrates relational modeling through perception, reasoning, and supervision levels.
    \item Extensive experiments on the OpenLaneV2 dataset validate the effectiveness of our approach, demonstrating multi-level relation benefit mutually. With sufficient improvements, our method surpasses previous methods across both perception and reasoning, setting a new state-of-the-art.
\end{itemize}

%% file: sections/2_related.tex
\section{Related Work}

\begin{figure*}[t!]
    \centering
    \includegraphics[width=0.98\linewidth]{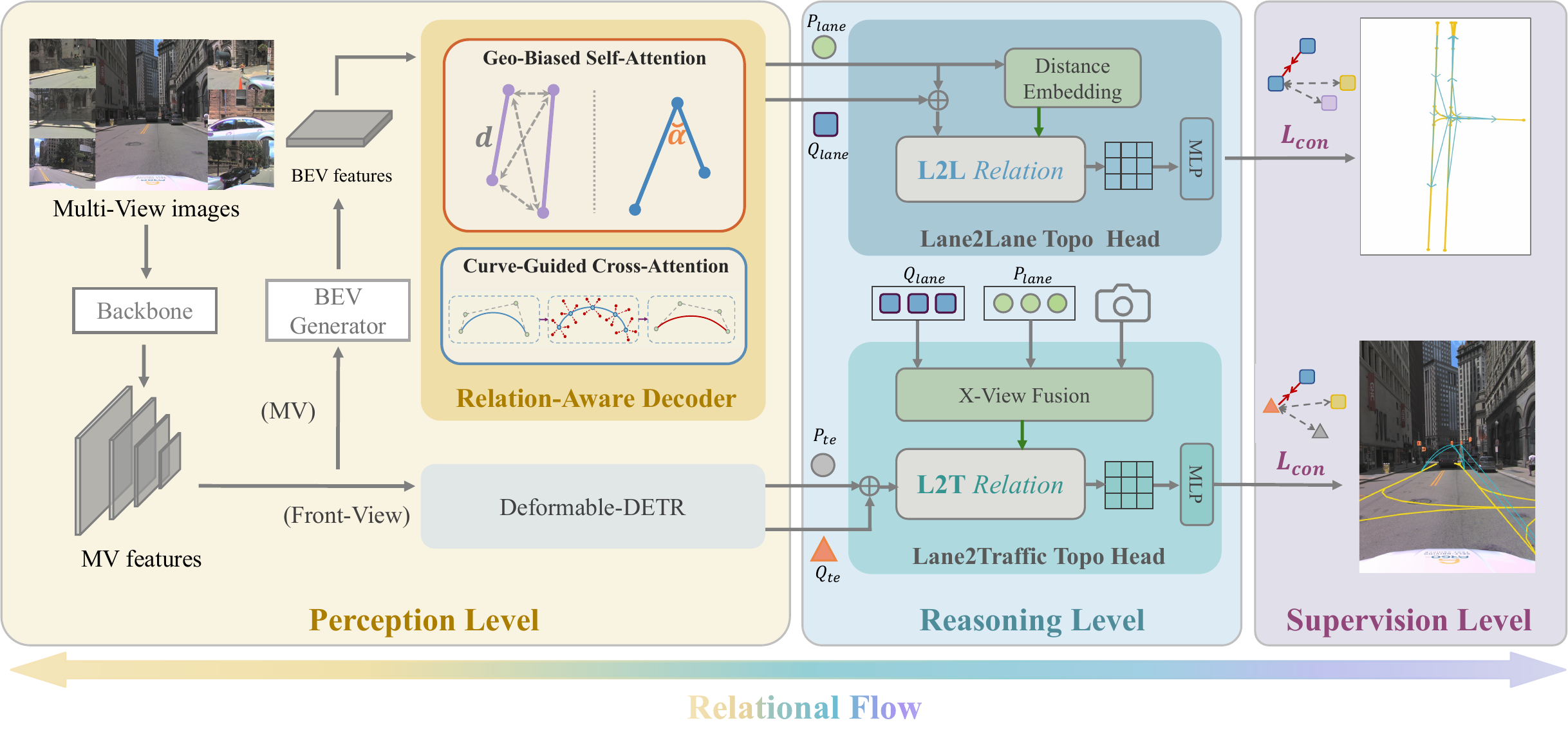}
    \vspace{-2mm}
    \caption{The overall framework of \modelname, 
    Our framework coherently integrates geometry-guided relational modeling across three levels:
    \textcolor{fig_per}{Perception Level:} A relation-aware lane decoder enrich lane features with geometric cues. \textcolor{fig_rea}{Reasoning Level:} Relation-enhanced topology heads infer topology between lanes (L2L) and between lanes and traffic elements (L2T) by embedding structural relations into relation embeddings,
    \textcolor{fig_sup}{Supervision Level:} Contrastive loss \textcolor{fig_sup}{$\mathcal{L}_{\text{con}}$} regularizes relational embeddings, pulling connected pairs closer and pushing apart non-connected pairs in the feature space.
    ~\includegraphics[height=0.78em]{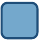} ($Q_{lane}$),~\includegraphics[height=0.78em]{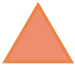} ($Q_{te}$) denote lane and traffic element queries, and ~\includegraphics[height=0.78em]{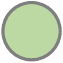} ($P_{lane}$) and~\includegraphics[height=0.78em]{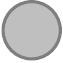} ($P_{te}$) denotes their predictions.}
    \label{fig:main}
    \vspace{2mm}
\end{figure*}

\subsection{3D Lane Detection}

3D lane detection provides the lane geometry for topology reasoning. Existing methods are typically designed for \textit{monocular} front-view images and fall into BEV-based and front-view-based paradigms. BEV-based methods employ inverse perspective mapping (IPM) to transform FV images into BEV space for lane prediction~\cite{3dlane, Genlanenet, 3dlanenet+, liu2022learning, Persformer, li2022reconstruct}.
However, IPM-based methods suffer from distortions on non-flat roads due to their planar assumption, limiting its reliability in real-world settings. 
Front-view-based approaches instead predict 3D lanes directly from FV features, avoiding IPM distortions. 
Recent works~\cite{Curveformer,Anchor3dlane,latr} adopt query-based detectors~\cite{detr, deformdetr} to model 3D lanes information end-to-end, achieving stronger performance.
Our work extends lane perception further by incorporating multi-view input and embedding relational cues directly into the perception process, enabling lane features to become inherently relation-aware.

\subsection{Online HD Map Construction}

Online HD map construction aims to dynamically generate structured map elements.
Early methods~\cite{Hdmapnet} use dense segmentation with heuristic vectorization.
VectorMapNet~\cite{Vectormapnet} advances this direction with an end-to-end detection-and-serialization pipeline.
More recent works~\cite{Maptr,Maptrv2,bemapnet,pivotnet,instagram,scalablemap,gemap,himap,admap} adopt Transformer-based architectures with curve- or point-based representations. MapTR~\cite{Maptr} introduces hierarchical query embeddings for polyline generation,
BeMapNet~\cite{bemapnet} and PivotNet~\cite{pivotnet} employ piecewise B\'ezier curves and dynamic points, and InstaGraM~\cite{instagram} formulates map element generation as a graph problem with GNNs.
GeMap~\cite{gemap} models structural and relational properties of map elements, but focuses mainly on individual geometries without explicitly capturing topology relationships among lanes, and it relies on polyline representations composed of equidistant points, which lack flexibility and precision~\cite{pivotnet,rastermap} for nuanced lane description. 
To improve computational efficiency,
various decoupled self-attention mechanisms~\cite{Maptrv2,admap,gemap} have been proposed for integrating intra-/inter-instance information.
However, topology reasoning among elements are limited in this area.

\subsection{Driving Scene Topology Reasoning}

Topology reasoning aims to capture the connectivity among road elements.
Early works like STSU~\cite{stsu} and TPLR~\cite{can2022topo} construct lane graphs directly in BEV space, but focus only on L2L reasoning and ignore interactions with traffic elements.
Recent methods~\cite{toponet,lanesegnet,topomlp,topo2d,roadpainter} begin to address both L2L and L2T reasoning. %
TopoNet~\cite{toponet} employs a GNN-based framework that enhances topology prediction through message passing between element embeddings. TopoMLP~\cite{topomlp} adopts MLP heads for more efficiency, and
LaneSegNet~\cite{lanesegnet}
proposes lane segment attention to capture intra-lane dependencies~\cite{Streammapnet}.
However, these methods do not explicitly model relationships among lanes or between lanes and traffic elements, limiting their ability to capture structural dependencies.
Topo2D~\cite{topo2d} incorporates additional 2D lane detection to support topology reasoning. TopoLogic~\cite{topologic} combines geometry (end-to-start) distance-based topology (GDT) estimation with query similarity-based relational modeling via GNNs, but applies geometric cues primarily as post-processing for L2L inference. Subsequently, works like\cite{topoformer,fu2025topopoint} move GDT into feature encoding, yet they still measures lane geometry relationships via only end-to-start patterns and rely on extra GNNs.

In contrast, our work takes a unified view: we embed geometric relational modeling coherently across perception, reasoning, and supervision, encoding comprehensive geometric priors (distance, angular relationships, spatial proximity) directly into feature learning for coherent joint optimization of perception and topology.

%% file: sections/3_method.tex
\section{Method}
\subsection{Overview}

\modelname weaves relational modeling through perception, reasoning, and supervision. Given multi-view images, the model comprises two branches: one decodes BEV lane features, and another detects front-view traffic elements. These outputs feed into relational topology heads to infer L2L and L2T connections. We inject relation cues early (in the decoder, \cref{sec:lc_decoder}), maintain them in reasoning (in the heads, \cref{sec:rel_reason}), and sharpen them through contrastive loss in supervision (\cref{sec:topo_loss}). This multi-level integration enables coherent perception and topology reasoning, where geometry guides both feature learning and relational inference in a unified framework. The following subsections detail each module.

\subsection{Relational Perception}
\label{sec:lc_decoder}
Lanes in real driving scenes exhibit strong geometric relationships, such as parallelism and convergence, that naturally guide perception but are often neglected in prior works. We embed these geometric relational cues directly into the decoder so that lane features become structurally informed. Concretely, we introduce a \textbf{relation-aware decoder} that comprises two complementary mechanisms: 1)  \textbf{geometry-biased self-attention} (\cref{sec:bias}) adds inter-lane geometric biases into the self-attention logits, steering each lane to attend to its peers in a relation-aware fashion; 2) \textbf{curve-guided cross-attention} (\cref{sec:ca}) samples points along the Bézier curve representation of each lane and performs deformable attention over those points. This enables context aggregation along the lane’s shape rather than just at control points. These two design choices ensure that lane queries evolve under structural guidance, improving robustness in downstream topology reasoning.

\subsubsection{Geometry-Biased Self-Attention}
\label{sec:bias}

Training queries in DETR-like decoders often suffer from slow convergence due to lack of structural inductive bias~\cite{relationdetr}. To address this, we encode pairwise lane geometry as an additional bias in self-attention. Given two lanes $\boldsymbol{l}_i$ and $\boldsymbol{l}_j$ with endpoints $\mathbf{p}^{s}_{\ast}, \mathbf{p}^{e}_{\ast}$, we define a symmetric \emph{minimum endpoint distance}:
\begin{equation}
\text{Dist}(\boldsymbol{l}_i,\boldsymbol{l}_j)
=
\min_{x\in\{s,e\},\, y\in\{s,e\}}
\left\lVert \mathbf{p}^{x}_{i}-\mathbf{p}^{y}_{j}\right\rVert_{2},
\label{eq:min_endpoint_dist}
\end{equation}
which measures the closest endpoint-to-endpoint proximity and is robust to endpoint ordering. We further compute an \emph{orientation difference} $\text{Angle}(\boldsymbol{l}_i,\boldsymbol{l}_j)$.
We concatenate these cues, and embed to biases via {a linear projection \texttt{GE}} (after sinusoidal 
encoding):
\begin{align}
    &\mathbf{Q} = \text{Softmax}\left(\frac{\mathbf{QK}^\top}{\sqrt{d_{\text{model}}}} + \textbf{Geometry}(\boldsymbol{l}, \boldsymbol{l})\right)\mathbf{V} \label{equ:biased_sa} \\
    &\text{\small $\textbf{Geometry}(\boldsymbol{l}, \boldsymbol{l})_{(i,j)} = $} \,{\texttt{GE}} \text{\small $\left( \text{Dist}(\boldsymbol{l}_i, \boldsymbol{l}_j) \| \text{Angle}(\boldsymbol{l}_i, \boldsymbol{l}_j) \right)$}.
\label{eq:sa}
\end{align}
We illustrate this mechanism in ~\includegraphics[height=0.9em]{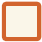} of~\cref{fig:lane_decoder} (a), where the \textbf{\textit{Geometry Bias}} represents the relation bias term $\textbf{Geometry}(\boldsymbol{l}, \boldsymbol{l})_{(i,j)}$ between $i$-th and $j$-th lanes. 
This mechanism enhances query learning by introducing structural bias~\cite{relationdetr}, while simultaneously capturing inherent geometric relationships between lanes.
Unlike Topologic~\cite{topologic}, which relies on GNNs to encode connectivity, our approach is simpler yet more effective: we directly encode geometric relationships as attention biases within self-attention, eliminating the need for separate graph propagation. Additionally, we incorporate angular information for comprehensive geometry relation modeling.

By enhancing geometrically proximal lanes, the model allocates greater focus to spatially relevant lanes through the interleaved attention process, leading to a more robust understanding of lane topology (detailed in~\cref{sec:l2l}). 
Our method differs from~\cite{relationdetr}, which encodes progressive cross-layer box positional relationships for individual objects.

\subsubsection{\Curve\ Cross-Attention}
\label{sec:ca}

\begin{figure}[t]
    \centering
    \includegraphics[width=0.95\linewidth]{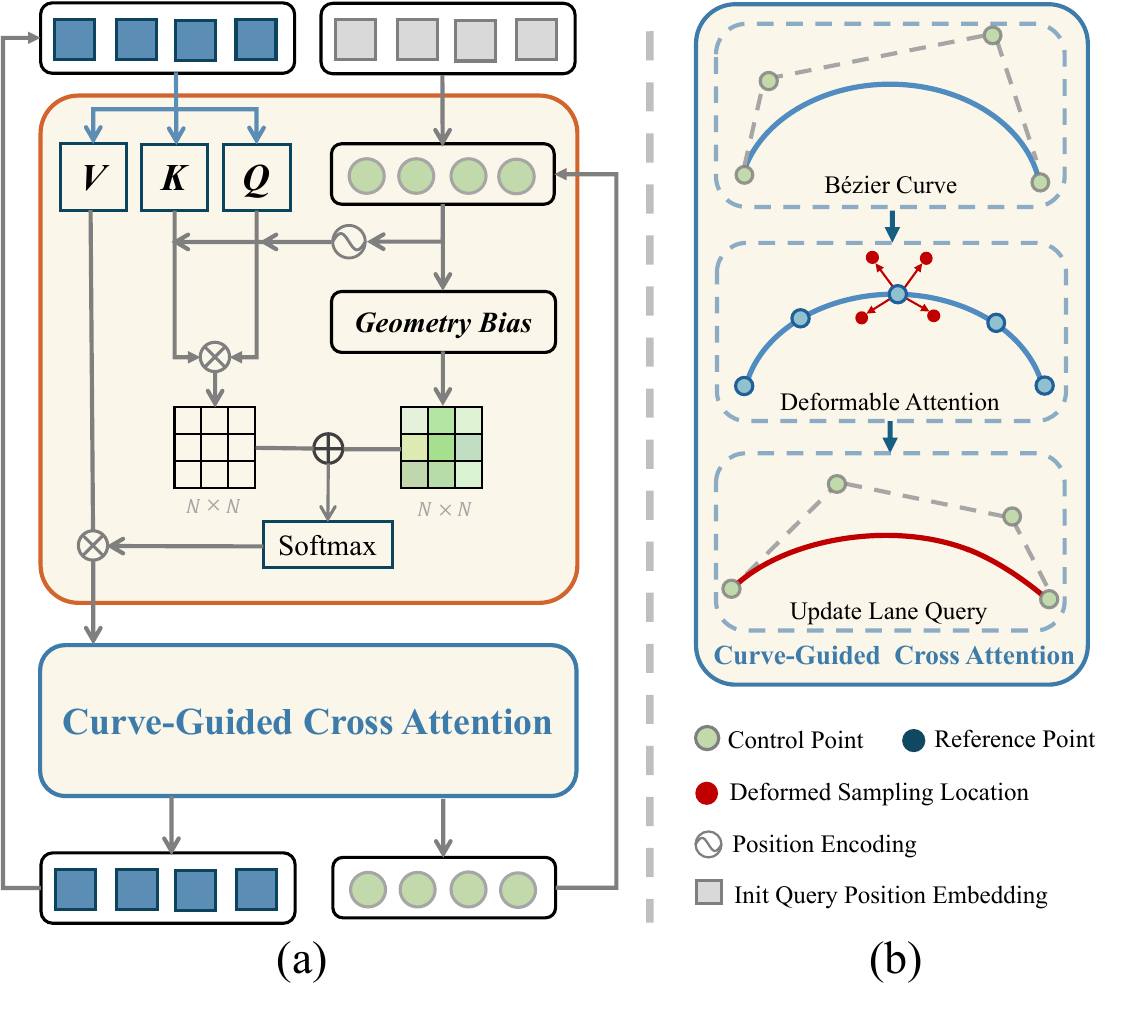}
    \caption{
    Illustration of our \bzr lane decoder layer, featuring our \underline{geometry-biased} SA~\includegraphics[height=0.9em]{figs/sa.pdf} and \underline{\curve} CA~\includegraphics[height=0.9em]{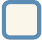}.}
    \label{fig:lane_decoder}
\end{figure}

Polyline representations with fixed spacing are inflexible~\cite{pivotnet,rastermap}. We instead adopt a compact cubic \bzr representation with four control points. However, two challenges arise: 1) the sparsity of control points limits feature aggregation; 2) intermediate control points may be misaligned with the underlying curve~\cref{fig:lane_decoder}(b). 

To address these issues, instead of relying solely on sparse control points as reference points in deformable attention as in TopoDBA~\cite{kalfaoglu2024topobda}, we sample $K$ points along the \bzr curve, which serves as reference points for feature aggregation (\cref{fig:lane_decoder} (b)).
Furthermore, to capture long-range intra-lane dependencies, we employ a shared lane query to integrally generate offsets and weights for all sampled $K$ points. This design enables contextual information flow holistically within each lane.

Given the feature map $\boldsymbol{x}$, the $l^{th}$ lane query $\boldsymbol{q}_l$ and its reference point $\boldsymbol{p}_l$, we adopt deformable attention mechanism~\cite{deformdetr} to update query features,  formulated as:
\begin{equation}
    \text{DeformAttn}(\boldsymbol{q}_l,\boldsymbol{p}_l,\boldsymbol{x}) = \sum^{M}_{m=1}\boldsymbol{W}_m \left[ \sum^{N}_{i=1}\sum^{K}_{j=1}A_{mlij}\cdot \boldsymbol{W}^{'}_m \boldsymbol{x}(\boldsymbol{p}_l + \Delta\boldsymbol{p}_{mlij})\right],
\end{equation}
where $M$ is the number of attention heads, $N$ denotes the number of offset locations per sampled point, and $K$ is the number of sampled points along each curve. $A_{mlij}$ and $\Delta\boldsymbol{p}_{mlij}$ denote the attention weights and sampling offset, respectively, for the $i^{th}$ offset of the $j^{th}$ sampled points along the curve in the $m^{th}$ attention head.

Unlike B\'{e}zierFormer~\cite{bezierformer}, which projects sampled points onto image feature maps and performs \verb|grid_sample|\footnote{This denotes the grid sampling operation defined in PyTorch.} to generate $N_{ref}$ point queries, each undergoing separate deformable cross-attention, 
our approach is simpler and more effective: it avoids the inefficiency of per-point queries and leverages global lane structure to guide local sampling, 
enhancing the modeling of long-range intra-lane dependencies.
Besides, unlike BeMapNet~\cite{bemapnet}, which represents each map element with multiple \bzr curves,
our single \bzr curve per lane balances efficiency and accuracy, as validated in~\cref{tab:bez}.

\subsection{Relational Reasoning}
\label{sec:rel_reason}
With relation-aware lane features in place, we infer connectivity in a unified relational space for both L2L and L2T. Our design departs from pipelines that either rely on brittle coordinate encodings~\cite{toponet,topomlp} or defer geometry to post-processing heuristics~\cite{topologic}. Instead, we embed relational priors \emph{inside} the pairwise embeddings and align cross-view features end-to-end, thereby reducing sensitivity to endpoint shifts and eliminating reliance on external fixes.

\subsubsection{Geometry-Enhanced L2L Reasoning}
\label{sec:l2l}

Existing methods~\cite{topomlp,roadpainter,topo2d} enhance lane features using coordinate-based encoding to predict L2L connectivity, which are, however, highly sensitive to endpoint shifts. Although~\cite{topologic} attempt to remedy this by incorporating endpoint geometric distance as extra topology features (GDT), but the post-hoc design of their GDT fails to internalize relational modeling in model learning (as we discussed in~\cref{sec:intro} and analysis in~\cref{sec:supp_exp_topologic}). This
leaves relational structure underutilized. 
Motivated by these limitations, we propose a geometry-enhanced L2L reasoning head that embeds geometric priors directly into pairwise embeddings.
Concretely, given refined lane features \(\mathbf{Q}_{\text{lane}} \in \mathbb{R}^{N\times C}\) and endpoints \(\mathbf{P}_{\text{lane}} \in \mathbb{R}^{N\times 2 \times 2}\), we first project \(\mathbf{Q}_{\text{lane}}\) into predecessor and successor embeddings with two MLPs (as L2L is directional, this choice is for capturing the asymmetric nature of L2L topology). Besides, we employ a sinusoidal positional encoding to produce lane positional embeddings based on $\mathbf{P}_{\text{lane}}$, yielding $\mathbf{PE}_{\text{lane}}$.
\begin{equation}
    \mathbf{G}_\text{L2L} = \left(\text{MLP}_{1}(\mathbf{Q}_{\text{lane}}) + \mathbf{PE}_{\text{pre}}\right) \, \mathbin{\copyright} \, \left(\text{MLP}_{2}(\mathbf{Q}_{\text{lane}})+ \mathbf{PE}_{\text{suc}}\right), 
    \label{eq:rel_gen}
\end{equation}

where $\mathbin{\copyright}$ denotes broadcast concatenation and $\mathbf{G}_\text{L2L} \in \mathbb{R}^{N \times N \times C}$.
To reinforce connectivity relationships and stabilize against endpoint noise, we introduce a distance embedding:
\begin{equation}
    \text{DistEmbed}_{\text{L2L}}^{i,j} = \text{MLP}(\texttt{distance}(\mathbf{P}_{i}^{e} - \mathbf{P}_{j}^{s})), \;\;\text{DistEmbed} \in \mathbb{R}^{N \times N \times C}
\end{equation}

where $\mathbf{P}_{i}^{e}$ and $\mathbf{P}_{j}^{s}$ denote the endpoint of lane $i$ and the starting point of lane $j$, respectively.
and then predict L2L topology via:
\begin{equation}
    \mathbf{T}_{\text{L2L}} = \text{MLP}(\mathbf{G}_\text{L2L} + \text{DistEmbed}_\text{L2L}), \;\;\mathbf{T}_{\text{L2L}} \in \mathbb{R}^{N \times N \times 1}\label{eq:rel_pred}
\end{equation}

By knitting geometry into the trainable relational embedding, our design internalizes robustness to small endpoint shifts and supports directional reasoning, unlike external post-processing hacks or pure coordinate encodings. This strengthens connected-pair discrimination and reduces false negatives in challenging scenarios. More details are illustrated in \textit{Appendix}~\cref{alg:geo_l2l}.
\begin{algorithm}
  \caption{Geometry-Enhanced L2L Reasoning}
  \label{alg:geo_l2l}
  \begin{algorithmic}[1]
    \Require Lane query features \(\mathbf{Q}_{\text{lane}} \in \mathbb{R}^{N \times C}\), lane endpoints \(\mathbf{P}_{\text{lane}} \in \mathbb{R}^{N \times 2 \times 2}\)
    \Ensure Connectivity logits \(\mathbf{T}_{\text{L2L}} \in \mathbb{R}^{N \times N \times 1}\)

    \State \(\mathbf{E}_{\text{pre}} \gets \mathrm{MLP}_1(\mathbf{Q}_{\text{lane}})\) \Comment{\( \in \mathbb{R}^{N \times \tfrac C2}\)}
    \State \(\mathbf{E}_{\text{suc}} \gets \mathrm{MLP}_2(\mathbf{Q}_{\text{lane}})\) \Comment{\(\in \mathbb{R}^{N \times \tfrac C2}\)}

    \State \((\mathbf{PE}_{\text{pre}}, \mathbf{PE}_{\text{suc}}) \gets \mathrm{PosEnc}(P_{\text{lane}})\)
    \Comment{sinusoidal positional encoding}

    \State \(\tilde{\mathbf{E}}_{\text{pre}} \gets \mathbf{E}_{\text{pre}} + \mathbf{PE}_{\text{pre}}\)
    \State \(\tilde{\mathbf{E}}_{\text{suc}} \gets \mathbf{E}_{\text{suc}} + \mathbf{PE}_{\text{suc}}\)

    \State \(\mathbf{G}_{\text{L2L}} \gets \mathrm{BroadcastConcat}(\tilde{\mathbf{E}}_{\text{pre}}, \tilde{\mathbf{E}}_{\text{suc}})\)
    \Comment{pairwise feature, \(\in \mathbb{R}^{N \times N \times C}\)}

    \State \(\mathrm{Dist}_{ij} \gets \texttt{distance}(P^{e}_{i} - P^{s}_{j})\)
    \Comment{distance: lane $i$ end point $\rightarrow$ lane $j$ start point, $\in \mathbb{R}^{N \times N \times 1}$}
    
    \State \(\mathrm{DistEmbed} \gets \mathrm{MLP}(\mathrm{Dist}_{ij})\) \Comment{\(\in \mathbb{R}^{N \times N \times C}\)}

    \State \(\mathbf{T}_{\text{L2L}} \gets \mathrm{MLP}(\mathbf{G}_{\text{L2L}} + \mathrm{DistEmbed})\)
    \Comment{\(\in \mathbb{R}^{N \times N \times 1}\)}

\Return \(\mathbf{T}_{\text{L2L}}\)
  \end{algorithmic}
\end{algorithm}

\subsubsection{X-View L2T Reasoning}
\label{sec:l2t}
L2T reasoning requires bridging two inherently different representations: BEV features for lanes and FV features for traffic elements. The disparity between these representations poses a challenge for learning spatially consistent relationships. Many prior efforts treat these features as if they lie in the same space and naïvely fuse or concatenate them~\cite{topomlp,roadpainter}. Topo2D~\cite{topo2d} mitigates this by injecting 2D priors via auxiliary decoders.

In contrast, we introduce a X-view fusion module that aligns BEV and FV features more directly and cohesively. First, we project predicted 3D lane coordinates \(\mathbf{P}_{\text{lane}}^{3D} \in \mathbb{R}^{N \times K \times 3}\) into the front-view plane to obtain \(\mathbf{P}_{\text{lane}}^{2D} \in \mathbb{R}^{N \times K \times 2}\). We then sample spatial features \(\mathbf{F}_{\text{lane}}^{2D}\) at those locations via \texttt{grid\_sample}. These sampled features serve as a bridge between BEV and FV spaces, aligned by 3D geometry.
Next, we fuse \(\mathbf{F}_{\text{lane}}^{2D}\) into lane queries along with their 2D positional embeddings (using similar sinusoidal encoding strategy as in L2L module), producing enhanced queries:
\begin{equation}
    \Tilde{\mathbf{Q}}_{\text{lane}} = \text{MLP}_1(\mathbf{Q}_{\text{lane}}) + \mathbf{F}_{\text{lane}}^{2D} + \mathbf{PE}_{\text{lane}}^{2D},
    ~~
    \Tilde{\mathbf{Q}}_{\text{te}} = \text{MLP}_2(\mathbf{Q}_{\text{te}}) + \mathbf{PE}_{\text{te}}^{2D}.
\end{equation}
Here the positional embeddings encode 2D spatial layout in the front-view, ensuring the fused features remain spatially coherent.
We then form the relational embedding \(\mathbf{G}_{\text{L2T}} \in \mathbb{R}^{N \times M \times C}\) by broadcast concatenation of \(\Tilde{\mathbf{Q}}_{\text{lane}}\) and \(\Tilde{\mathbf{Q}}_{\text{te}}\), and feed it into an MLP to predict the L2T topology:
\begin{equation}
\mathbf{T}_{\text{L2T}} = \text{MLP}(\mathbf{G}_{\text{L2T}}), \;\;\mathbf{T}_{\text{L2T}} \in \mathbb{R}^{N \times M \times 1}.
\end{equation}
By aligning BEV lane features to FV contexts via projected geometry and position embeddings, we reduce misalignment between two space representation, enabling more robust lane–traffic associations. Empirically, this design lowers erroneous associations in setups where naïve fusion would fail, especially at larger distances or oblique angles (shown in the qualitative comparison in~\cref{sec:supp_qua_res}).

\begin{table*}[tp!]
\centering
\caption{\textbf{Performance comparison with existing methods} on the OpenLane-V2 \texttt{subsetA} and \texttt{subsetB} dataset under the latest V2.1.0 evaluation. Results for RoadPainter$^\ddagger$ derive from their paper which uses old metrics (due to the absence of open-source code or model). TopoMLP$^{\dagger}$ results were obtained with their provided model, and TopoFormer$^\S$ denotes our reimplemented~\cite{topoformer} results, due to their $\times$2 input size, another fine-tuning stage and lack of runnable code.
Metrics include detection accuracy (DET$_l$ and DET$_t$) and topology reasoning accuracy (TOP$_{ll}$ and TOP$_{lt}$), with overall score (OLS) indicating aggregated performance.}
\label{table:sota_a}
\small
\resizebox{0.95\textwidth}{!}{
\setlength\tabcolsep{5pt}
\renewcommand\arraystretch{0.9}
\begin{tabular}{c|l|c|cc|ccccc}
\toprule
\noalign{\smallskip}
\textbf{Subset} & \textbf{Method} & \textbf{Venue} & \textbf{Backbone} & \textbf{Epoch} & \textbf{DET$_l$} $\uparrow$ & \textbf{DET$_t$} $\uparrow$ & \textbf{TOP$_{ll}$} $\uparrow$ & \textbf{TOP$_{lt}$} $\uparrow$ & \textbf{OLS} $\uparrow$ \\ 
\noalign{\smallskip}
\hline
\noalign{\smallskip}
\multicolumn{1}{>{\centering}p{10pt}|}{\cellcolor{white}\multirow{9}{*}{\parbox{10pt}{\centering{\texttt{setA}}}}} & VectorMapNet & ICML2023 & ResNet-50 & 24 & 11.1 & 41.7 & 2.7 & 9.2 & 24.9 \\
& MapTR & ICLR2023 & ResNet-50 & 24 & 17.7 & 43.5 & 5.9 & 15.1 & 31.0 \\
& TopoNet & Arxiv2023 & ResNet-50 & 24 & 28.6 & 48.6 & 10.9 & 23.8 & 39.8\\
& TopoMLP$^{\dagger}$ & ICLR2024 & ResNet-50 & 24 & 28.5 & \underline{50.5} & 21.7 & \underline{27.3} & \underline{44.5} \\
& RoadPainter$^\ddagger$  & ECCV2024 & ResNet-50 & 24 & \underline{30.7} & 47.7 & 7.9 & 24.3 & 38.9 \\
& TopoLogic & NeurIPS2024 & ResNet-50 & 24 & 29.9 & 47.2 & \underline{23.9} & 25.4 & 44.1 \\
& TopoFormer$^\S$ & CVPR2025 & ResNet-50 & 24 & 31.5 & 47.8 & 22.6 & 26.8 & 44.7 \\
\cline{2-10}
\noalign{\smallskip}
& \multicolumn{1}{c|}{\cellcolor[gray]{0.92}Ours} & \cellcolor[gray]{0.92}- & \cellcolor[gray]{0.92}ResNet-50 & \cellcolor[gray]{0.92}24 & \cellcolor[gray]{0.92}\textbf{33.8} & \cellcolor[gray]{0.92}\textbf{50.9 }& \cellcolor[gray]{0.92}\textbf{29.2} & \cellcolor[gray]{0.92}\textbf{32.2} & \cellcolor[gray]{0.92}\textbf{48.9} \\
& \multicolumn{1}{c|}{\cellcolor[gray]{0.92}\textcolor{mgreenpp}{\textit{Improvement}}} & \cellcolor[gray]{0.92}- & \cellcolor[gray]{0.92}- & \cellcolor[gray]{0.92}- & \cellcolor[gray]{0.92}\textcolor{mgreenpp}{3.1 $\uparrow$} & \cellcolor[gray]{0.92}\textcolor{mgreenpp}{0.4 $\uparrow$} & \cellcolor[gray]{0.92}\textcolor{mgreenpp}{5.3 $\uparrow$} & \cellcolor[gray]{0.92}\textcolor{mgreenpp}{4.9 $\uparrow$} & \cellcolor[gray]{0.92}\textcolor{mgreenpp}{4.4 $\uparrow$} \\
\cmidrule[0.6pt]{1-10}
\multicolumn{1}{>{\centering}p{10pt}|}{\cellcolor{white}\multirow{7}{*}{\parbox{10pt}{\centering{\texttt{setB}}}}} & TopoNet & Arxiv2023 & ResNet-50 & 24 & 24.3 & 55.0 & 6.7 & 16.7 & 36.8 \\
& TopoMLP$^\dagger$ & ICLR2024 & ResNet-50 & 24 & 26.0 & \underline{58.2} & 21.0 & \underline{19.8} & \underline{43.6} \\
& RoadPainter$^\ddagger$  & ECCV2024 & ResNet-50 & 24 & \underline{28.7} & 54.8 & 8.5 & 17.2 & 38.5 \\
& TopoLogic & NeurIPS2024 & ResNet-50 & 24 & 25.9 & 54.7 & \underline{21.6} & 17.9 & 42.3 \\
& TopoFormer$^\S$ & CVPR2025 & ResNet-50 & 24 & 29.2 & 55.1 & 20.9 & 19.7 & \underline{43.6} \\
\cline{2-10}
\noalign{\smallskip}
& \multicolumn{1}{c|}{\cellcolor[gray]{0.92}Ours} & \cellcolor[gray]{0.92}- & \cellcolor[gray]{0.92}ResNet-50 & \cellcolor[gray]{0.92}24 & \cellcolor[gray]{0.92}\textbf{32.6} & \cellcolor[gray]{0.92}\textbf{58.8} & \cellcolor[gray]{0.92}\textbf{31.8} & \cellcolor[gray]{0.92}\textbf{25.8} & \cellcolor[gray]{0.92}\textbf{49.7} \\
& \multicolumn{1}{c|}{\cellcolor[gray]{0.92}\textcolor{mgreenpp}{\textit{Improvement}}} & \cellcolor[gray]{0.92}- & \cellcolor[gray]{0.92}- & \cellcolor[gray]{0.92}- & \cellcolor[gray]{0.92}\textcolor{mgreenpp}{3.4 $\uparrow$} & \cellcolor[gray]{0.92}\textcolor{mgreenpp}{0.6 $\uparrow$} & \cellcolor[gray]{0.92}\textcolor{mgreenpp}{10.2 $\uparrow$} & \cellcolor[gray]{0.92}\textcolor{mgreenpp}{6.0 $\uparrow$} & \cellcolor[gray]{0.92}\textcolor{mgreenpp}{6.1 $\uparrow$} \\
\noalign{\smallskip}
\bottomrule
\end{tabular}
}
\vspace{3mm}
\end{table*}

\subsection{Loss Functions}

\subsubsection{Perception Loss}

For lane detection, we employ Focal Loss~\cite{focalloss} for classification and a combination of point-wise L1 loss and Chamfer distance loss over $K$ sampling points to guide accurate geometric regression.
For traffic element detection, we again use Focal Loss~\cite{focalloss} for classification, along with L1 loss and GIoU loss~\cite{giouloss} to supervise bounding box regression over $(x,y,w,h)$.

\subsubsection{Relational Supervision}
\label{sec:topo_loss}
Most existing methods rely solely on Focal Loss to determine connectivity over pairs.
However, this approach primarily emphasizes binary classification without explicitly distinguishing the relative importance of connected and non-connected pairs.
To better capture topological relationships, we augment this with a \textbf{contrastive} InfoNCE loss~\cite{infonce} over pairwise relation embeddings.

Concretely, consider the L2T case (the same logic applies to L2L). Let $N$ be the number of lane queries and $M$ the number of traffic element queries. The relational head produces an embedding tensor $\mathbf{G}_{\text{L2T}} \in \mathbb{R}^{N \times M \times C}$ and outputs a logit tensor $\mathbf{T}_{\text{L2T}}$.
We treat pairs labeled as “connected” as positives, and all others as negatives. To sharpen discrimination, we introduce a \textbf{hard negative mining} strategy: for each positive pair, we select the top-$n$ hardest negative pairs according to predicted logits, focusing the contrastive signal where confusion is highest.
We then apply a symmetric InfoNCE loss (in the spirit of CLIP~\cite{clip}-style contrastive formulations) over these logits. Let $\mathbf{v}^+$ and $\{\mathbf{v}^-\}$ denote the positive and selected negative logits, then:
\begin{equation}
\label{eq:nce1}
\begin{aligned}
\mathcal{L}_{\text{con}} &= -\log \frac{\exp(\mathbf{v}^+)}{\exp(\mathbf{v}^+) + \sum_{\mathbf{v}^-}\exp(\mathbf{v}^-)} 
 = \log \left[1 + \sum_{\mathbf{v}^-}\exp(\mathbf{v}^- - \mathbf{v}^+)\right],
\end{aligned}
\end{equation}
To handle multiple positives for a given anchor, we extend~\Cref{eq:nce1}, following previous works~\cite{idol,wang2023making}, to:
\begin{equation} \label{eq:nce2}
 \mathcal{L}_{\text{con}} = \log \left[1 + \sum_{\mathbf{v}^+}\sum_{\mathbf{v}^-}\exp(\mathbf{v}^- - \mathbf{v}^+)\right].
\end{equation}

%% file: sections/4_exp.tex
\section{Experimental Results}
\subsection{Dataset and Metrics}

\noindent\textbf{Dataset.} We evaluate our method on OpenLane-V2~\cite{Openlane-v2}, a large-scale dataset specifically designed for topology reasoning in autonomous driving 
comprising two subsets: \texttt{subsetA} (from Argoverse-V2~\cite{argoverse2}) and \texttt{subsetB} (from nuScenes~\cite{nuscenes}).

\noindent\textbf{Evaluation Metrics.} Following the official evaluation of OpenLane-V2~\cite{Openlane-v2}, we utilize DET$_l$ and DET$_t$ to measure detection accuracy for lanes and traffic elements, respectively. For topology reasoning, we employ TOP$_{ll}$ and TOP$_{lt}$ to assess L2L and L2T relationship prediction. The overall performance is quantified using the OpenLane-V2 Score (OLS):
\begin{equation}
    \text{OLS} = \frac{1}{4}\left[\text{DET}_l + \text{DET}_t + f(\text{TOP}_{ll}) + f(\text{TOP}_{lt})\right],
\end{equation}
where $f$ denotes the square root function.
Our evaluations follow the latest version (\verb|V2.1.0|) of the metrics, as updated in the official OpenLane-V2 GitHub repository\footnote{\url{https://github.com/OpenDriveLab/OpenLane-V2/issues/76}}.

\subsection{Implementation Details}
\noindent\textbf{Model Details.} We use a ResNet-50 backbone to extract features, coupled with a pyramid network, FPN, for multi-scale feature learning.
Following prior work~\cite{topologic,toponet}, a BEVFormer encoder~\cite{Bevformer} with 3 layers is employed to generate a BEV feature map of size $100\times 200$.
We employ six decoder layers, using 300 queries for the lane decoder and 100 queries for the traffic element decoder following~\cite{topomlp}.

\noindent\textbf{Training Details.} We utilize the AdamW optimizer~\cite{adamw} for model training, with a weight decay of 0.01 and an initial learning rate of $2.0\times10^{-4}$, which decays following a cosine annealing schedule.
Training is conducted for 24 epochs using a total batch size of 8 on 8 NVIDIA 4090 GPUs. Input images are resized to 1024$\times$800, following~\cite{topomlp}. Due to space limitations, the training loss is provided in the \textit{Appendix}~\cref{sec:supp_loss}.

\begin{table*}[h!]
\vspace{2mm}
    \centering
    \small
    \caption{\textbf{Ablation study on our  key components}: \textcolor{fig_per}{{Relational Perception}} (geometry-biased SA and curve-guided CA), \textcolor{fig_rea}{{Relational Reasoning}}, and \textcolor{fig_sup}{{Relational Supervision}}. 
    Colored rows mark the milestones upon equipping the respective modules.}
    \label{tab:combined_ablation}
    \resizebox{0.99\textwidth}{!}{
    \begin{tabular}{c cc cc c ccccc cccc}
        \toprule
        & \multicolumn{2}{c}{\textcolor{fig_per}{\textbf{Rel. Perception}}} 
        & \multicolumn{2}{c}{\textcolor{fig_rea}{\textbf{Rel. Reasoning}}} 
        & \multicolumn{1}{c}{\textcolor{fig_sup}{\textbf{Rel. Sup.}}} 
        & \multicolumn{5}{c}{\textbf{Metrics}} 
        & \multicolumn{4}{c}{\textbf{Efficiency \& Performance}} \\[0.5ex]
        
        \cmidrule(lr){2-3} \cmidrule(lr){4-5} \cmidrule(lr){6-6} \cmidrule(lr){7-11} \cmidrule(lr){12-15}
        
        \#L & SA & CA & L2L & L2T & NCE & DET$_l$ & DET$_t$ & TOP$_{ll}$ & TOP$_{lt}$ & OLS & Size \scriptsize{(MB)} & Lat. \scriptsize{(ms)} & $\Delta$ Cost & $\Delta$ OLS\\
        \midrule
        
        \textcolor{grayres}{\#1} & & & & & & \textcolor{grayres}{27.7} & \textcolor{grayres}{50.1} & \textcolor{grayres}{21.8} & \textcolor{grayres}{28.4} & \textcolor{grayres}{44.5} 
        & \textcolor{grayres}{218.7} & \textcolor{grayres}{159.7} & \textcolor{grayres}{--} & \textcolor{grayres}{--} \\
        \textcolor{grayres}{\#2} & \textcolor{fig_per}{$\checkmark$} & & & & & \textcolor{grayres}{28.1} & \textcolor{grayres}{48.8} & \textcolor{grayres}{24.7} & \textcolor{grayres}{28.3} & \textcolor{grayres}{44.9} 
        & \textcolor{grayres}{218.7} & \textcolor{grayres}{160.6} & \textcolor{grayres}{+0.6\%} & \textcolor{grayres}{+0.4} \\
        \rowcolor{tab_per}
        \textcolor{fig_per}{\#3} & \textcolor{fig_per}{$\checkmark$} & 
        \textcolor{fig_per}{$\checkmark$} & & & & 32.7 & 50.1 & 26.8 & 29.9 & 47.3 
        & 223.2 & 161.4 & +1.1\% & +2.8
        \\
        
        \textcolor{grayres}{\#4} & \textcolor{fig_per}{$\checkmark$} & \textcolor{fig_per}{$\checkmark$} & \textcolor{fig_rea}{$\checkmark$} & & & \textcolor{grayres}{33.7} & \textcolor{grayres}{48.0} & \textcolor{grayres}{28.5} & \textcolor{grayres}{29.7} & \textcolor{grayres}{47.4} 
        & \textcolor{grayres}{224.2} & \textcolor{grayres}{161.8} & \textcolor{grayres}{+1.3\%} & \textcolor{grayres}{+2.9}
        \\
        \textcolor{grayres}{\#5} & \textcolor{fig_per}{$\checkmark$} & \textcolor{fig_per}{$\checkmark$} & & \textcolor{fig_rea}{$\checkmark$} & & \textcolor{grayres}{33.5} & \textcolor{grayres}{48.4} & \textcolor{grayres}{26.4} & \textcolor{grayres}{31.4} & \textcolor{grayres}{47.4} 
        & \textcolor{grayres}{224.2} & \textcolor{grayres}{161.7} & \textcolor{grayres}{+1.3\%}  & \textcolor{grayres}{+2.9} \\
        
        \rowcolor{tab_rea}
        \textcolor{fig_rea}{\#6} & \textcolor{fig_per}{$\checkmark$} & \textcolor{fig_per}{$\checkmark$} & \textcolor{fig_rea}{$\checkmark$} & \textcolor{fig_rea}{$\checkmark$} & & {33.9} & 50.3 & 28.6 & 30.6 & 48.2 
        & 224.8 & 162.2 & +1.6\%  & +3.7 \\
        
        \rowcolor{tab_sup}
        \textcolor{fig_sup}{\#7} & \textcolor{fig_per}{$\checkmark$} & \textcolor{fig_per}{$\checkmark$} & \textcolor{fig_rea}{$\checkmark$} & \textcolor{fig_rea}{$\checkmark$} & \textcolor{fig_sup}{$\checkmark$} & \textbf{33.8} & \textbf{50.9} & \textbf{29.2} & \textbf{32.2} & \textbf{48.9} & {224.8} & {162.2} & {+1.6\%} & {+4.4} \\
        \bottomrule
    \end{tabular}
    }
\vspace{2mm}
\end{table*}

\subsection{Main Results}

We compare our model against SOTA methods on OpenLane-V2, with results shown in~\cref{table:sota_a}. On \texttt{subsetA}, \modelname achieves an OSL of 48.9, surpassing all previous methods by a significant margin (+4.4). The gains are consistent gains across metrics, with particularly strong improvements in lane detection (+3.1 DET$_l$) and topology (+5.3 TOP$_{ll}$, +4.4 TOP$_{lt}$).
On \texttt{subsetB}, \modelname further improves to 49.7 OLS (+6.1), driven by large improvements in topology reasoning (+10.2 TOP$_{ll}$ and +6.0 TOP$_{lt}$), while maintaining consistent improvements in detection metrics. These results confirm that our relational modeling at perception, reasoning, and supervision levels systematically strengthens both perception and topology reasoning. 
Although our model adopts the same traffic decoder as prior work~\cite{topomlp}, it still achieves +0.4/+0.6 gains in DET$_t$ on \texttt{setA}/\texttt{setB}, suggesting that relational cues learned for L2T reasoning also benefit traffic element perception.

\begin{figure*}[!tbp]
    \centering
   \includegraphics[width=\textwidth,height=0.6\textheight]{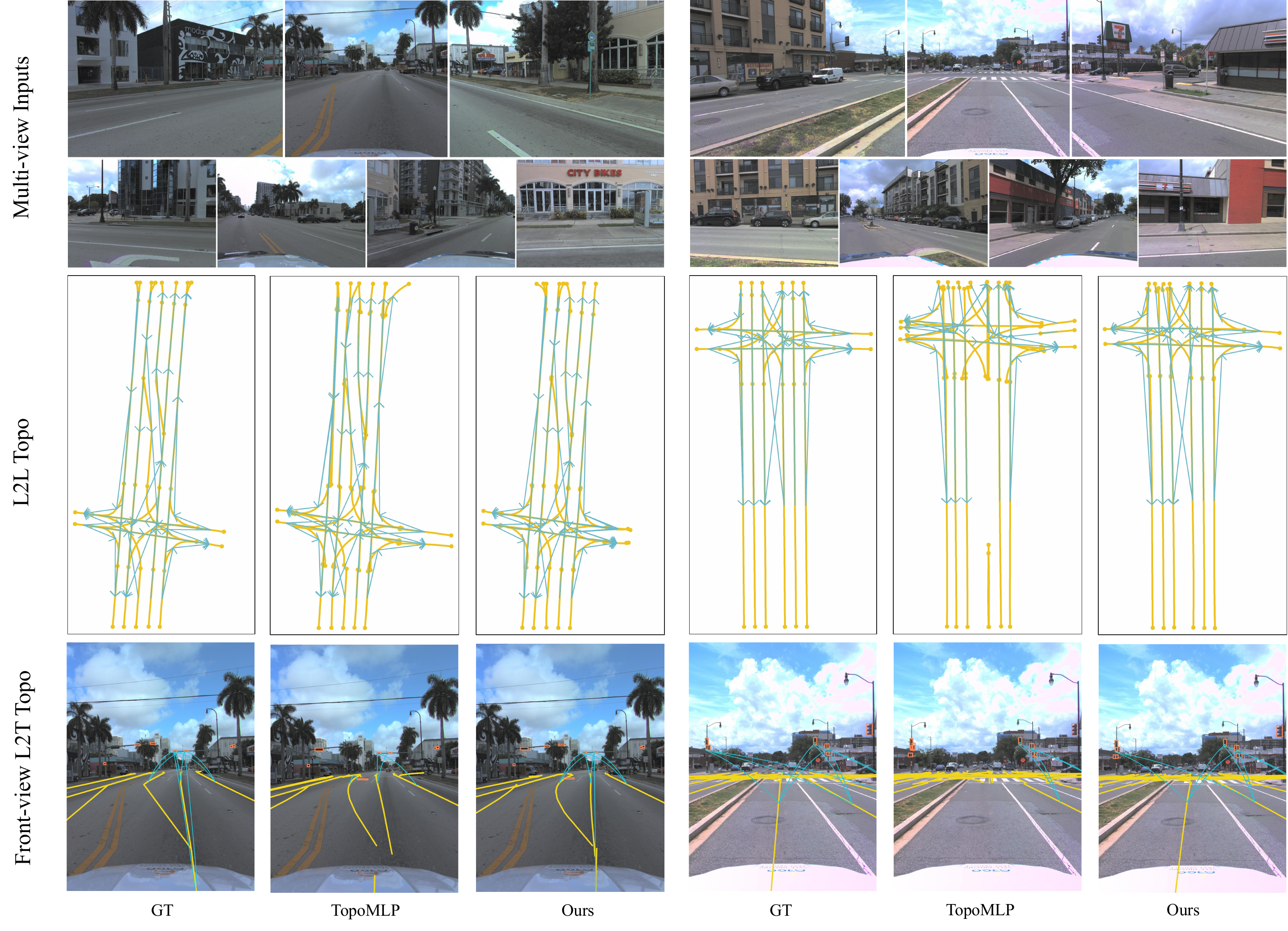}
   \vspace{1mm}
    \caption{Qualitative results comparison. The 1$^\text{st}$ row presents the multi-view input images, the {2$^\text{rd}$ row} illustrates the predicted L2L topology in BEV, where {light blue} \textcolor{mblue}{arrowlines} indicate directed L2L \textcolor{mblue}{connectivity} estimation (detailed lane centerline predictions are provided in the supplementary material). The {final row} depicts the front-view L2T topology predictions, where {orange} \textcolor{mgreen}{boxes} highlight detected \textcolor{mgreen}{traffic elements} (\eg, traffic lights and signs), {yellow} \textcolor{myellow}{lines} denote projected \textcolor{myellow}{lane centerlines}, and {blue} lines represent lane-to-traffic element (L2T) associations. }
    \label{fig:vis_supp1}
    \vspace{3mm}
\end{figure*}

Overall, these results highlight the superiority of \modelname, which sets a new state-of-the-art on both OpenLane-V2 \texttt{SubsetA} and \texttt{SubsetB}.
To further illustrate its effectiveness, we present \textbf{qualitative results} in~\cref{fig:teaser,fig:vis_supp1}, where our model produces more accurate lane predictions, better-aligned connection points and more accurate topology estimation in complex driving scenes. More detailed visualizations are provided in \textit{Appendix}~\cref{sec:supp_qua_res}.

\subsection{Ablation Studies}

To thoroughly evaluate our core designs, we conduct ablation studies on OpenLane-V2 \texttt{subsetA}. Our baseline model (\#1) adopts a deformable-DETR decoder following~\cite{topomlp} with lightweight MLP-based topology heads. Results are shown in~\cref{tab:combined_ablation}.

Additional analyses are provided in our \textit{Appendix}, covering: comparison of our geometry-biased \texttt{SA} and L2L relational modeling against previous methods~\cite{topologic,topoformer,fu2025topopoint} (\cref{sec:supp_exp_topologic}). We also discuss the limitations of our method and potential directions for future work.

\noindent\textbf{Effect of Relational Perception.}
Our relation-aware lane decoder includes Geometry-Biased Self-Attention (\texttt{SA}) and \Curve\ Cross-Attention (\texttt{CA}).
Replacing standard self-attention with \texttt{SA} (\#2) improves TOP$_{ll}$ by +2.9, showing the benefit of encoding inter-lane geometry.
Besides, we compare our geometry encoding with the distance topology method from Topologic in \textit{Appendix}~\cref{tab:compare_logic_main}, further validating the advantages of our method.
Introducing \texttt{CA} (\#3) further boosts DET$_l$ by +4.6, as curve-guided sampling better aggregates long-range features. 
Together, \texttt{SA} and \texttt{CA} (\#3) deliver substantial gains over baseline (\#1): +5.0 DET$_l$, +5.0 TOP$_{ll}$, +1.5 TOP$_{lt}$, and +2.8 OLS. 
These results confirm the combined benefits within our relational perception.

\noindent\textbf{Effect of Relational Reasoning.}
We next evaluate our relational topology heads. 
Adding our geometry-enhanced L2L head (\#3\(\rightarrow\)\#4) improves TOP$_{ll}$ by +1.7 and also raises DET$_t$ by +1.0, validating its effectiveness in capturing L2L relationships and showing that stronger lane connectivity reasoning indirectly benefits perception. 
Replacing the baseline L2T head with our cross-view design (\#3\(\rightarrow\)\#5) improves TOP$_{lt}$ by +1.5, demonstrating better lane–traffic alignment. 
Integrating both heads (\#6) yields complementary gains (+1.8 TOP$_{ll}$, +0.7 TOP$_{lt}$, +0.9 OLS), confirming that relational reasoning at both L2L and L2T levels jointly strengthens performance.

\noindent\textbf{Effect of Relational Supervision:}  
Finally, we incorporate our contrastive loss for additional supervision in topology learning. Compared to \#6, adding our contrastive regularization (\#7) provides further gains (+0.6 TOP$_{ll}$, +1.6 TOP$_{lt}$), validating that relational supervision helps structure the embedding space for more discriminative topology learning.

\noindent{\textbf{Efficiency and Performance.}}
{To quantify the overhead of our relational modules, we report model size and inference latency for all ablation variants in~\cref{tab:combined_ablation} on a single RTX~4090 GPU, as shown in the ``Efficiency \& Performance" part.%
Equipping our relational perception and relational reasoning increases latency only slightly, from 159.7\,ms to 162.2\,ms (about $\approx$1.6\% over the baseline), while yielding a substantial OLS gain of +4.4 points (\#1~$\rightarrow$~\#7). Note that our relational supervision is used only during training (\ie, \#6~$\rightarrow$~\#7 introduces no extra inference overhead). Thanks to the lightweight and effective design of our multi-level relational modeling, these gains come at negligible computational cost.}

\begin{table}[h]
\vspace{2mm}
\centering
\caption{\textbf{Comparison with Geometric Distance Topology (GDT).} "-- GDT$_{\text{L2L}}$": removes GDT from L2L inference. "$^\dagger$": inference using the hyperparameters learned from training.
"+ GDT$_{\text{lane}}$": replaces our geometry encoding with GDT for lane representation. 
"+ GDT$_{\text{L2L}}$": ensembles our L2L predictions with GDT at inference.}
\label{tab:compare_logic_main}
\setlength{\tabcolsep}{5pt} 
\renewcommand{\arraystretch}{1.0}
\begin{tabular}{l|cccc|c}
    \toprule
    \noalign{\vskip -0.3mm}
     Model & DET$_l$ & DET$_{t}$ & TOP$_{ll}$ & TOP$_{lt}$ & OLS \\
    \hline
    \noalign{\smallskip}
    Topologic & 29.9 & 47.2 & 23.9 & 25.4 & 44.1 \\
    Topologic -- {GDT}$_{\text{L2L}}$ & 29.9 & 47.2 & 11.6 & 25.4 & 40.4 \\
    Topologic$^\dagger$ & 29.9 & 47.2 & 23.2 & 25.4 & 43.9 \\
    \hline
    \noalign{\smallskip}
    Ours & \textbf{33.8} & \textbf{50.9} & \textbf{29.2} & \textbf{32.2} & \textbf{48.9} \\
    Ours + {GDT}$_{\text{lane}}$ & 32.8 & 50.0 & 28.4 & 30.8 & 47.9 \\
    Ours + {GDT}$_{\text{L2L}}$ &33.8 & 50.9 & 28.7 & 32.2 & 48.7 \\
    \bottomrule
\end{tabular}
\vspace{2mm}
\end{table}

\noindent\textbf{Comparison with Geometric Distance Topology (GDT).} TopoLogic~\cite{topologic} proposes a geometric distance topology (GDT)%
approach that shares certain similarities with our method, but differs fundamentally in design and effectiveness. 
\textbf{First}, for lane representation learning, GDT-based methods~\cite{topologic,topoformer,fu2025topopoint} only captures end-to-start point distances between lanes, while our geometry-biased self-attention encodes richer spatial cues including inter-lane distances and angular relations (\eg, parallelism, perpendicularity), which are common in real-world road layouts.
\textbf{Second}, for L2L topology reasoning, TopoLogic applies GDT as a post-hoc refinement \textbf{only} at inference time with manually tuned hyperparameters, whereas our method integrates geometric cues directly into L2L relation features during end-to-end training.
As shown in Table~\ref{tab:compare_logic_main},
when TopoLogic uses the GDT hyperparameters learned during training, its performance drops.
Further replacing our geometry encoding with GDT leads to performance drops (DET$_l$: 33.8$\to$32.8, TOP$_{ll}$: 29.2$\to$28.4), and applying GDT to our L2L inference also degrades TOP$_{ll}$ (29.2$\to$28.7), highlighting the superiority of our learnable geometric modeling approach.
More detailed analysis can be found in our appendix material.

\vspace{2mm}
\subsection{Comparsion of B\'{e}zier representation.}
\label{sec:supp_bzr}
We compare our B\'{e}zier-based lane representation with other alternative formulations in~\cref{tab:bez},
B\'{e}zierFormer~\cite{bezierformer} uses \texttt{grid\_sample} to sample features and separately predicts offsets and attention weights for each point within deformable attention, which increases computational overhead and compromises instance-level consistency in lane prediction.
In contrast, our method predicts offsets and attention weights jointly for the entire lane instance using a single lane query, resulting in improved efficiency and coherence.
BeMapNet~\cite{bemapnet} employs a piecewise B\'{e}zier representation by dynamically predicting the number of segments, which increases model complexity and uncertainty.
In the task of lane topology reasoning, such dynamic segmentation does not yield performance gains and lacks the structural regularity that our fixed B\'{e}zier formulation provides.

\begin{table}[h!]
\centering
\small
\caption{Comparsion with different b\'{e}zier representation.}
\label{tab:bez}
\setlength{\tabcolsep}{8pt} 
\renewcommand{\arraystretch}{1.0}
\begin{tabular}{l|cccc|c}
    \toprule
     
     Lane Rep. & DET$_l$ & DET$_{t}$ & TOP$_{ll}$ & TOP$_{lt}$ & OLS \\
    \hline
    \addlinespace[1pt]
    BeMapNet & 29.9 & 50.0 & 25.9 & 28.9 & 46.2 \\
    B\'{e}zierFormer & 31.0 & 49.7 & 26.1 & 29.6 & 46.5 \\
    Ours & \textbf{33.8} & \textbf{50.9} & \textbf{29.2} & \textbf{32.2} & \textbf{48.9} \\
    \bottomrule
\end{tabular}
\end{table}

\section{Conclusion}
We present \modelname~, a unified framework that coherently integrates geometric relational modeling across perception, reasoning, and supervision for driving scene topology reasoning.
By embedding comprehensive geometric priors into relation-aware lane features and topology heads, \modelname enables end-to-end joint optimization where perception and reasoning mutually reinforce each other.
Extensive experiments on OpenLane-V2 demonstrate significant improvements, establishing new state-of-the-art performance. We believe \modelname demonstrates the effectiveness of coherent relational modeling for driving scene perception.

\begin{acks}
This work was supported in part by Guangdong S\&T Programme with Grant No. 2024B0101030002, the Basic Research Project No. HZQB-KCZYZ-2021067 of Hetao Shenzhen-HK S\&T Cooperation Zone, the Shenzhen Outstanding Talents Training Fund 202002, the NSFC with Grant No. 62293482, by NSFC with Grant No. 62573371,  by the Guangdong Province Radio Science Data Center with grant No. 2025B1212070001, by the Shenzhen General Program \,\,\,No. JCYJ20220530143600001, by the Guangdong Research Project No. 2017ZT07X152 and No. 2019CX01X104, by the Guangdong Provincial Key Laboratory of Future Networks of Intelligence (Grant No. 2022B1212010001), by the Guangdong Provincial Key Laboratory of BigData Computing CHUK-Shenzhen, by the NSFC 61931024 \& 12326610, by the Shenzhen Key Laboratory of Big Data and Artificial Intelligence (Grant No. SYSPG20241211173853027), by National Key Research and Development Program of China: 2025YFF0515300 and 2025YFF0515304, the Open Project Program (Grant No. QHSF-CS-2606) of Key Laboratory of Tibetan Information Processing, Ministry of Education, by the Shenzhen-Hong Kong Joint Funding No. SGDX20211123112401002, and by Tencent \& Huawei Open Fund 2024E0009\&202301030019.
\end{acks}